\documentclass{article}

\usepackage[preprint]{neurips_2022}

\usepackage[utf8]{inputenc} 
\usepackage[T1]{fontenc}    
\usepackage{hyperref}       
\usepackage{url}            
\usepackage{booktabs}       
\usepackage{amsfonts}       
\usepackage{nicefrac}       
\usepackage{microtype}      
\usepackage{xcolor}         
\usepackage{todonotes}
\usepackage{amsmath}
\usepackage{subcaption}
\usepackage{caption}
\usepackage{multirow}
\usepackage{cases}
\usepackage{enumitem}
\setlist[itemize]{align=parleft,left=0pt..1em}

\usepackage{siunitx}

\graphicspath{{figs/}}

\newcommand{\rpm}{\raisebox{.2ex}{$\scriptstyle\pm$}}

\setcitestyle{numbers,square}

\usepackage{amsthm}
\newtheorem{theorem}{Theorem}[section]
\newtheorem{lemma}[theorem]{Lemma}
\newtheorem{corollary}[theorem]{Corollary}

\title{Asynchronous Neural Networks\\for Learning in Graphs}

\author{%
Lukas Faber\\
ETH Zurich, Switzerland\\
\texttt{lfaber@ethz.ch}\\
\And
Roger Wattenhofer\\
ETH Zurich, Switzerland\\
\texttt{wattenhofer@ethz.ch}\\
}

\begin{document}
\maketitle

\begin{abstract}
This paper studies asynchronous message passing (AMP), a new paradigm for applying neural network based learning to graphs. Existing graph neural networks use the synchronous distributed computing model and aggregate their neighbors in each round, which causes problems such as oversmoothing and limits their expressiveness. On the other hand, AMP is based on the \textit{asynchronous} model, where nodes react to messages of their neighbors individually. We prove  that (i) AMP can simulate synchronous GNNs and  that (ii) AMP can theoretically distinguish any pair of graphs. We experimentally validate AMP's expressiveness. Further, we show that AMP might be better suited to propagate messages over large distances in graphs and performs well on several graph classification benchmarks.
\end{abstract}

\section{Introduction}
Graph Neural Networks (GNNs) have become the de-facto standard model for applying neural networks to graphs in many domains~\citep{bian2020rumor, gilmer2017neural, hamilton2017inductive, jumper2021highly, kipf2017semi, velivckovic2018graph, wu2020comprehensive}. Internally, nodes in GNNs use message passing, they communicate with their neighboring nodes for multiple synchronous rounds. We believe that the way GNNs realize this communication is not ideal. In GNNs, all nodes speak concurrently, and a node does not listen to individual neighbors but only to an aggregated message of all neighbors. 
In contrast, humans politely listen when a neighbor speaks, then decide whether the information was relevant, and what information to pass on. 

\begin{figure}[h]
    \centering
    \includegraphics[width=0.65\textwidth]{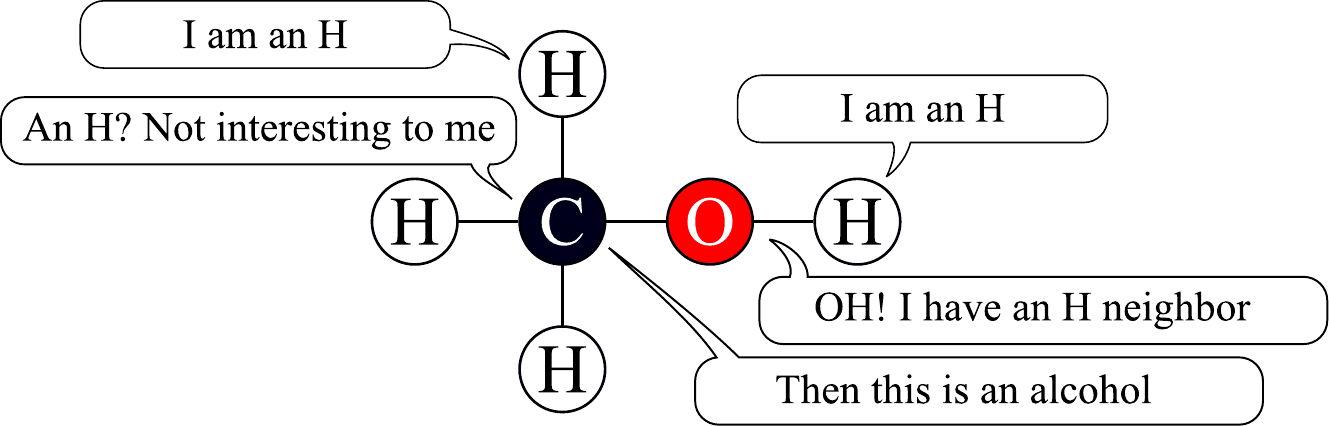}
    \caption{Detection of an alcohol (a C atom with an OH group) with AMP. H atoms send a message to their neighbors. Every node can choose to ignore the message or react to it. The C atom is not interested in H neighbors and discards the message. On the other hand, the O atom reacts and sends a message on its own. This message is now relevant to the C atom.}
    \label{fig:comic}
\end{figure}

The way humans communicate is in line with the asynchronous communication model~\citep{peleg2000distributed}. 
In the asynchronous model, nodes do not communicate concurrently. In fact, a node only acts initially or when it
receives a message. If a node receives a new message from one of its neighbors, it updates its state, and then potentially sends a message on it own. This allows nodes to listen to individual neighbors and not only to aggregations. Figure~\ref{fig:comic} illustrates how this interaction can play out.

We make the following contributions.
\begin{itemize}
    \item We introduce AMP, a new paradigm for learning neural architectures in graphs. Instead of nodes acting synchronously in rounds, nodes in AMP interact asynchronously by exchanging and reacting to individual messages.
    \item We theoretically examine the expressive power of AMP. Our two main results are that \textit{in principle} (i) AMPs can simulate GNNs and that (ii) AMP with message delays can go beyond all Weisfeiler-Lehman tests~\citep{shervashidze2011weisfeiler} and solve any graph isomorphism problem.
    \item We examine how AMP can transmit information to far-away nodes. Since AMP handles messages individually and is not limited by the number of communication rounds, AMP can combat the underreaching, oversmoothing, and oversquashing problems that traditional GNNs encounter when propagating information over long distances (many layers).
    \item We run experiments on (i) established GNN expressiveness benchmarks to demonstrate that AMP outperforms all existing methods in distinguishing graphs beyond the $1-$WL algorithm. We introduce (ii) synthetic datasets to show that AMP is well suited to propagate information over large distances. Finally, we study (iii) established graph classification benchmarks to show that AMP performs comparably to existing GNNs in classification accuracy.
\end{itemize}

\section{Related Work}
Virtually all GNNs follow the synchronous message passing framework of distributed computing, first suggested by \citet{gilmer2017neural} and \citet{battaglia2018relational}. The underlying idea is that nodes have an embedding and operate in rounds. In each round, every node computes a message and passes the message to every adjacent node. Then, every node aggregates the messages it receives and uses the aggregation to update its embedding. There exist variations of this framework. For example, edges can also have embeddings, or one can add a global sharing node to allow far away nodes to directly share information~\citep{battaglia2018relational}. Following the initial work of \citet{scarselli2008graph}, different implementations for the individual steps in the message passing framework exist, e.g.,~\citep{brody2022attentive, hamilton2017inductive, kipf2017semi, niepert2016learning, velivckovic2018graph, xu2018representation, xu2019powerful}. However, these GNN architectures all experience common problems:

\textbf{Oversmoothing.} A problem that quickly emerged with GNNs is that we cannot have many GNN layers~\citep{li2019deepgcns, li2018deeper}. Each layer averages and hence smooths the neighborhood information and the node's features. This effect leads to features converging after some layers~\citep{oono2020graph}, which is known as the oversmoothing problem. Several works address the oversmoothing problem, for example by sampling nodes and edges to use in message passing~\citep{feng2020graph, hasanzadeh2020bayesian, rong2020dropedge}, leveraging skip connections~\citep{chen2020simple, xu2018representation}, or additional regularization terms~\citep{chen2020measuring, zhao2020pairnorm, zhou2020towards}. Thanks to its asynchrony, AMP does not average over neighborhood messages and is not exposed to the oversmoothing problem.

\textbf{Underreaching.} Using normal GNN layers, a GNN with $k$ layers only learns about nodes at most $k$ hops away. A node cannot act correctly if it would need information that is $k+1$ hops away. This problem is called underreaching~\citep{barcelo2020logical}. There exist counter measures, for example, having a global exchange of features~\citep{gilmer2017neural, wu2021representing} or spreading information using diffusion processes~\citep{gasteiger2019diffusion, scarselli2008graph}. Methods that help against oversmoothing are usually also applied against underreaching, since we can use more layers and increase the neighborhood size. In AMP, because of asynchrony, some nodes can be involved in the computation/communication much more often than others; this helps AMP to gather information from further away, which is a countermeasure against underreaching.

\textbf{Oversquashing.} In many graphs, the size of $k-$hop neighborhoods grows substantially with $k$. This requires squashing more and more information into a node embedding of static size. Eventually this leads to the congestion problem (too much information having to pass through a bottleneck) that is well known in distributed computing (e.g.~\citep{sarma2012distributed}) and goes by the name of oversquashing for GNNs \citep{alon2021on, topping2022understanding}. One approach to solve oversquashing is introducing additional edges that function as shortcuts to non-direct neighbors~\citep{bruel2022rewiring}. Dropping-based methods~\citep{feng2020graph, hasanzadeh2020bayesian, rong2020dropedge} that help against oversmoothing can also reduce oversquashing by reducing the size of the neighborhoods. AMP is again naturally more resilient than synchronous GNNs, since information can be transferred in a more guided fashion.

\textbf{$1-$WL Limit.} \citet{xu2019powerful} and \citet{morris2019weisfeiler} show that GNNs are limited in their expressiveness by the $1-$Weisfeiler-Lehman test ($1-$WL), a heuristic algorithm to evaluate graph isomorphism~\citep{shervashidze2011weisfeiler}. Hoever, there exist simple structures that the $1-$WL test cannot distinguish that we want to detect with GNNs~\citep{garg2020generalization}. Therefore, several augmentations to GNNs exist that include additional features, such as ports, IDs, or angles between edges for chemistry datasets~\citep{gasteiger2020directional, loukas20graph, sato2019approximation, sato2021random}, run multiple rounds over slight perturbations of the same graph~\citep{bevilacqua2022equivariant, papp2021dropgnn, vignac2020building}, or use higher-order information~\citep{chen2019equivalence, maron2019provably, morris2019weisfeiler}. Since in AMP nodes do not act at the same time, they can be distinguished more easily. We show that AMP can handle $1-$WL and higher order WL, in principle even graph isomorphism.

We believe these four classic GNN problems have their root cause in the \textit{synchronous aggregation} of (almost) all neighbors. The aggregation smooths the neighborhood features, and nodes are exposed to all the information in their increasing $k-hop$ neighborhood. This limits the number of rounds, causing underreaching. If nodes with identical neighbors act synchronously, they stay identical and within $1-$WL expressiveness. 

There are two main approaches for distributed computing \citep{peleg2000distributed, wattenhofer2020mastering}. In the synchronous approach, all nodes operate in synchronous rounds, and in each round each node sends a message to every neighbor. \citet{sato2019approximation} and \citet{loukas20graph} show that current graph neural networks using message passing as in \citet{gilmer2017neural} follow this approach. The antipodal paradigm in distributed computing is the asynchronous model \citet{peleg2000distributed}. In the asynchronous model, nodes do not act at the same time. Instead nodes act on receiving a single message. If receiving a message, a node may change its internal state and potentially create new messages itself. Currently, all GNN approaches follow the synchronous message passing model. In this paper, we want to explore the efficacy of asynchronous message passing.  So we ask a natural question: Is it possible, and is it worthwhile for GNNs to handle messages asynchronously and individually?

\section{The Asynchronous Model and its Challenges}
We illustrate the dynamics in AMP in Figures \ref{fig:comic} and \ref{fig:node_recurrent}. Normally, AMP picks one node and sends it a special initial message to start the computation which looks as follows: A node is in a state $h$. Upon receiving a message $m$, a node can react to it and change its internal state. The node can then decide to emit a message $m'$ itself based on its (potentially) new state. The node does not know how many and which other neighbor nodes will receive the message, therefore, $m'$ does not contain information about the receiver. If we look at the dynamics of one node in isolation, the behavior mimics that of sequence to sequence models~\citep{sutskever2014sequence} that have seen great popularity and success in Natural Language Processing. In principle, every node receives a sequence of messages that the node can use to create new internal states and produce a sequence of messages. The node does not need to produce a message in every step. For example, in Figure~\ref{fig:node_recurrent} the node sends no message in step $i+1$. Compared to NLP sequence to sequence models, the asynchronous messaging passing must overcome two additional challenges:

\textbf{Unknown input sequences.} In most sequence-to-sequence language models, we want to transform a known input sequence into an output sequence. In NLP, input sequences are often sentences and the sequence elements are words, for example, when we want to translate sentences. In such cases, the input sequence is entirely known at the start and sequence elements do not depend on each other. When a node $v$ in AMP receives a message $m_0$ it is not clear what the message $m_1$ will be nor do we know which node will send it. Even worse, node $v$ might emit a message $m_0'$ that influences the message $m_1$ node $v$ will receive later. This may happen since the outputs of node $v$ are the inputs of its neighboring nodes. This mutual dependence between inputs and outputs makes learning harder.

\textbf{Partial information available.}
In natural language processing, the input sequence contains all the needed information to produce the target prediction. Consider again the translation setting, where sentences are inputs and words are sequence elements. Every word has access to the entire input sequence and thus all the information it needs. However, most nodes in graphs do not have access to all information they need. Figure~\ref{fig:needed_propagation} illustrates the problem with a simple example. The node on the right $v$ needs to know if there is a red node in the graph. There is one such node, $s$. However, in between those two nodes are nodes $t$ and $u$. For both $t$ and $u$, the red node $s$ is not important since $t$ and $u$ are only interested in the number of blue nodes. Therefore, if either $t$ or $u$ decide to not forward information about the red node, the information that $s$ exists never reaches $v$ and $v$ cannot classify correctly. The takeaway from this example is that nodes not only need to understand what messages are important to them and how to use these messages, but nodes also need to identify the messages that are important to other nodes and propagate these messages.

\begin{minipage}{0.57\textwidth}
    \centering
    \def\svgwidth{\textwidth}
\begingroup%
  \makeatletter%
  \providecommand\color[2][]{%
    \errmessage{(Inkscape) Color is used for the text in Inkscape, but the package 'color.sty' is not loaded}%
    \renewcommand\color[2][]{}%
  }%
  \providecommand\transparent[1]{%
    \errmessage{(Inkscape) Transparency is used (non-zero) for the text in Inkscape, but the package 'transparent.sty' is not loaded}%
    \renewcommand\transparent[1]{}%
  }%
  \providecommand\rotatebox[2]{#2}%
  \newcommand*\fsize{\dimexpr\f@size pt\relax}%
  \newcommand*\lineheight[1]{\fontsize{\fsize}{#1\fsize}\selectfont}%
  \ifx\svgwidth\undefined%
    \setlength{\unitlength}{389.94480819bp}%
    \ifx\svgscale\undefined%
      \relax%
    \else%
      \setlength{\unitlength}{\unitlength * \real{\svgscale}}%
    \fi%
  \else%
    \setlength{\unitlength}{\svgwidth}%
  \fi%
  \global\let\svgwidth\undefined%
  \global\let\svgscale\undefined%
  \makeatother%
  \begin{picture}(1,0.27576538)%
    \lineheight{1}%
    \setlength\tabcolsep{0pt}%
    \put(0,0){\includegraphics[width=\unitlength,page=1]{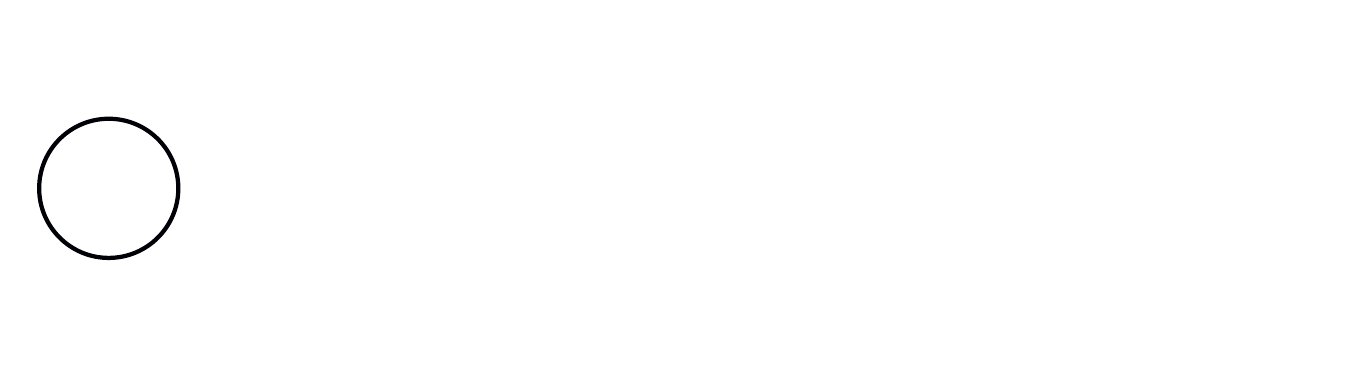}}%
    \put(0.04,0.128){\color[rgb]{0,0,0}\makebox(0,0)[lt]{\lineheight{1.25}\smash{\begin{tabular}[t]{l}$h_{i-1}$\end{tabular}}}}%
    \put(0,0){\includegraphics[width=\unitlength,page=2]{recurrentfig.pdf}}%
    \put(0.1,0.229){\color[rgb]{0,0,0}\makebox(0,0)[lt]{\lineheight{1.25}\smash{\begin{tabular}[t]{l}$m_{i-1}$\end{tabular}}}}%
    \put(0.1,0.04){\color[rgb]{0,0,0}\makebox(0,0)[lt]{\lineheight{1.25}\smash{\begin{tabular}[t]{l}$m_{i-1}'$\end{tabular}}}}%
    \put(0,0){\includegraphics[width=\unitlength,page=3]{recurrentfig.pdf}}%
    \put(0.293,0.128){\color[rgb]{0,0,0}\makebox(0,0)[lt]{\lineheight{1.25}\smash{\begin{tabular}[t]{l}$h_i$\end{tabular}}}}%
    \put(0,0){\includegraphics[width=\unitlength,page=4]{recurrentfig.pdf}}%
    \put(0.335,0.229){\color[rgb]{0,0,0}\makebox(0,0)[lt]{\lineheight{1.25}\smash{\begin{tabular}[t]{l}$m_i$\end{tabular}}}}%
    \put(0.335,0.04){\color[rgb]{0,0,0}\makebox(0,0)[lt]{\lineheight{1.25}\smash{\begin{tabular}[t]{l}$m_i'$\end{tabular}}}}%
    \put(0,0){\includegraphics[width=\unitlength,page=5]{recurrentfig.pdf}}%
    \put(0.505,0.128){\color[rgb]{0,0,0}\makebox(0,0)[lt]{\lineheight{1.25}\smash{\begin{tabular}[t]{l}$h_{i+1}$\end{tabular}}}}%
    \put(0,0){\includegraphics[width=\unitlength,page=6]{recurrentfig.pdf}}%
    \put(0.568,0.229){\color[rgb]{0,0,0}\makebox(0,0)[lt]{\lineheight{1.25}\smash{\begin{tabular}[t]{l}$m_{i+1}$\end{tabular}}}}%
    \put(0,0){\includegraphics[width=\unitlength,page=7]{recurrentfig.pdf}}%
    \put(0.734,0.128){\color[rgb]{0,0,0}\makebox(0,0)[lt]{\lineheight{1.25}\smash{\begin{tabular}[t]{l}$h_{i+2}$\end{tabular}}}}%
    \put(0,0){\includegraphics[width=\unitlength,page=8]{recurrentfig.pdf}}%
    \put(0.8,0.229){\color[rgb]{0,0,0}\makebox(0,0)[lt]{\lineheight{1.25}\smash{\begin{tabular}[t]{l}$m_{i+2}$\end{tabular}}}}%
    \put(0.8,0.04){\color[rgb]{0,0,0}\makebox(0,0)[lt]{\lineheight{1.25}\smash{\begin{tabular}[t]{l}$m_{i+2}'$\end{tabular}}}}%
    \put(0,0){\includegraphics[width=\unitlength,page=9]{recurrentfig.pdf}}%
  \end{picture}%
\endgroup%

    \captionof{figure}{AMP dynamics for one node in isolation. The node receives a sequence of messages which it can use to update its state and (possibly, not always, e.g. $h_{i+1}$) emit a sequence of messages.}
    \label{fig:node_recurrent}
\end{minipage}\hfill%
\begin{minipage}{0.36\textwidth}
    \centering
    \includegraphics[width=0.7\textwidth]{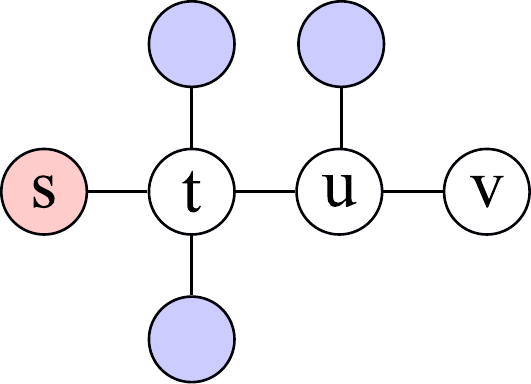}
    \captionof{figure}{Node $v$ needs a message from $s$ but can only receive it if both $t$ and $u$, which are not interested in $s$, forward the message.}
    \label{fig:needed_propagation}
\end{minipage}%

\textbf{Long-Range Information Propagation.} We now discuss how to propagate information in AMP, in particular over long distances. To send information to a node far away, many messages must be sent and received. This leads to many rollouts of the node update function. We know from recurrent neural networks that many rollouts can lead to slow and unstable training~\citep{hochreiter1997long}. Therefore, we want to discuss several possible designs for the node update function.

\textbf{Recurrent Layers.} One straightforward idea is to use layers that we know from recurrent neural networks. For example, we can use a simple feedforward layer as in recurrent neural networks (we call this architecture AMP-RNN). Gating is a useful concept in RNNs, therefore, we create variants of AMP that use the GRU~\citep{cho2014learning} (AMP-GRU) or the LSTM~\citep{hochreiter1997long} (AMP-LSTM). While gating allows a node to minimize the change in state for a non-relevant message, neither of these ideas allows a node to truly terminate. We will later show that this causes problems when we want to extrapolate the learned AMP variants to much larger graphs.

\textbf{Self-supervised termination.} Therefore, we also study ideas for self-supervised termination in RNNs. \citet{graves2016adaptive} introduce the idea of adaptive computation time (ACT). With ACT, a RNN can learn how many steps it wants to perform for each sequence element. Before every step, the model learns a halting probability and finishes when the sum of probabilities is sufficiently close to $1$. AMP-ACT adopts this idea and computes halting probabilites for every message. If the sum of these halting probabilities is large enough, the node stops updating. Its final state is a probability-weighted average of all states. We experiment with another version, AMP-Iter, that is inspired by the IterGNN work of \citet{tang2020towards}. In IterGNN, nodes also compute a halting probability on every message. However, IterGNNs combine these probabilities multiplicatively instead of additively.

\section{Expressiveness of AMP}
\subsection{AMP is at least as expressive as GNNs}
First, we examine the relationship between AMP and message passing GNNs. Our first main theoretical result is that AMP can simulate the execution of a GNN, showing that AMP is at least as powerful as GNNs. The proof idea follows the so-called $\alpha$ synchronizer~\citep{awerbuch1985complexity}. The core concepts of the $\alpha$ synchronizer are \emph{pulses} and \emph{safe} nodes. Nodes emit a pulse when they start simulating the next synchronous round. A node is safe if every message it sent in the current pulse was received and processed by its neighbors. Nodes determine safety using acknowledgments messages. If a node and all its neighbors are safe, the node proceeds to the next pulse.

For concreteness, we will simulate a simple version of GIN~\citep{xu2019powerful} that we call sGIN. sGIN uses the identity as message function, and nodes compute their new state by simply summing their neighbors' states, applying a linear layer on the sum and a ReLu activation afterward. Without loss of generality, we assume that every layer in sGIN has the same dimension $d$ and that sGIN has a total of $L$ layers.

Our variant of AMP that mimics sGIN has states of size $d+3$. The additional three entries store information required for the simulation: AMP stores the number of neighbors $w$ for which the node is waiting for a message at the current pulse, the number of nodes $u$ that are still unsafe at the current pulse, and the number of layers $l$ left to simulate. We assume that all nodes in AMP know their degree $D$.\footnote{This assumption simplifies the proof but does not provide additional information. Every node in AMP could also learn its degree by flooding the network once and storing the count in a state that is never modified again.} Every node starts in the state $(x, 0, 0, L)$, where $x$ are the input features of sGIN.

Nodes in AMP always send their old state $s$ plus $3$ additional bits for coordination as messages. The first two bits are the \texttt{pulse} bit to signal a new pulse and the \texttt{safe} bit to inform neighbors that the node became safe. Furthermore, we need an \texttt{origin} bit that is only set on the very first message which requires special handling. In every message, exactly one of these bits is set.

Let a node have the state $(s, w, u, l)$ and receive a message $(m, \texttt{pulse}, \texttt{safe}, \texttt{origin})$. Table~\ref{tab:f_table_simulation} describes the function $f$ that computes an updated node state and a new message to send in return. The symbol $\bot$ denotes that no message is sent.

\begin{table*}[ht]
\centering
\resizebox{0.6\textwidth}{!}{
\begin{tabular}{@{}l*{6}{S[table-format=-3.4]}@{}}
\toprule
{Condition} & {s'} & {w'} & {u'} & {l'} & {message}\\\midrule
\makebox{l=0} & {s} & {w} & {u} & {l} & {$\bot$}\\
{\texttt{origin}=1} & {s} & {D-1} & {D} & l & {\texttt{pulse}} \\
{u=0} & {m} & {D-2} & {D} & {l} & {\texttt{pulse}} \\
{\texttt{safe}=1} & {s} & {w} & {u-1} & {l} & {$\bot$} \\
{w=0} & {ReLu(u(s+m))} & {0} & {u-1} & {l-1} & {\texttt{safe}} \\
{\texttt{pulse}=1} & {s+m} & {w-1} & {u} & {l} & {$\bot$} \\\bottomrule
\end{tabular}}
\caption{Update function $f$ used by every node in AMP to simulate a synchronous GNN.}
\label{tab:f_table_simulation}
\end{table*}

\begin{theorem}AMP can simulate sGIN.
\end{theorem}
\label{theorem:simulation}
\begin{proof}
We prove by induction that when a node sends a new pulse, the node and its neighbors finished the previous pulse with the same state as sGIN. The statement  holds at the beginning since nodes in sGIN and AMP start with the same features.

Let us assume the statement holds and node $v$ is about to emit a new pulse. Nodes emit pulses when they are in the third condition (the second condition is a special case for the very first message). Before $v$ can send another pulse, its state entry $u$ needs to become $0$. For this, two things need to happen: $v$ needs to receive a \texttt{safe} message from every neighbor (condition 4) and $v$ needs to become safe itself. $v$ becomes safe if it receives a message from every neighbor (condition 5) and updates the sum of these messages the same way as GIN. Note that this definition is slightly different from \citet{awerbuch1985complexity}: A node is safe when it processed every received message, not when every sent message was acknowledged. These two scenarios are equivalent. Thus, when $v$ becomes safe, it has the same state as in sGIN before the next layer, the same is true for $v$'s neighbors.

Note that the universal approximation theorem~\citep{royden1988real} tells us that we can model $f$ to an arbitrarily close approximation using a neural network with at least two layers and non-linear activations and sufficient depth. Because $f$ is a simple function of cascading ifs, we can exactly model $f$ using three layers with a $ReLu$ activation. For space reasons, we defer this construction to Appendix~\ref{sec:simulation_weights}.
\end{proof}

We can easily extend this construction to more complex GNN variants. For example, we can divide $(u(s+m))$ by $D$ to use \texttt{mean} aggregation. We can also accumulate messages in separate features from the node state and allow the update to depend on both state and aggregated messages. Another variation is to normalize messages by the sender's and receiver's degree to emulate GCN~\citep{kipf2017semi}.

\subsection{AMP with random message delays}
To understand the expressiveness in more detail, we need to distinguish two variants of AMP. In the first variant of AMP, we assume that message have a random delay before they arrive. For examples delays could be chosen uniformly in the interval [0,1]. Delays can act as a source of randomness, which makes this variant very powerful.
\begin{lemma}
AMP with random message delays can create IDs for every node in a star graph in $O(k\cdot 6^{k})$ time, where $k$ is the degree of the center node.
\end{lemma}
\label{lemma:identifiers}

We proof Lemma~\ref{lemma:identifiers} in the Appendix. It is possible to generalize the star setting to general graphs. We send an arbitrary node an initial message that will start to distribute IDs in its induced star graph neighborhood. After this node assigns an ID to every neighbor, it makes the node with the next-highest ID the new center node who assigns IDs next. Some neighbors of the new center might already have an ID, only the ones without participate.

\begin{corollary}
AMP with random message delays can create unique node identifiers in every graph.
\end{corollary}
\label{corollary:identifiers}

\citet{loukas20graph} prove that a GNN with unique identifiers for every node, sufficiently wide and powerful layers, and sufficient depth is Turing complete. Theorem~\ref{corollary:identifiers} allows us to build identifiers for every node even if there are none. Now, we can simulate a Turing complete GNN according to Theorem~\ref{theorem:simulation}. This yields our main result on AMP's expressiveness.
\begin{theorem}
AMP with random message delays can distinguish any pair of graphs.
\end{theorem}

\subsection{AMP with constant message delays.}
\begin{figure}
    \begin{subfigure}[t]{0.3\textwidth}
        \centering
        \includegraphics[width=\textwidth]{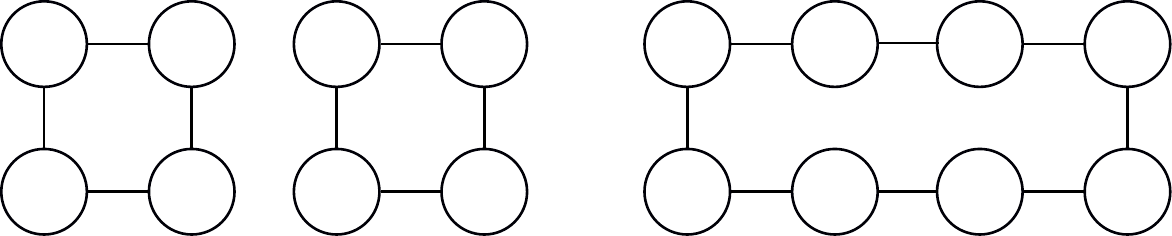}
        \caption{}
        \label{fig:beyond1wl_garg}
    \end{subfigure}\hfill%
    \begin{subfigure}[t]{0.3\textwidth}
        \centering
        \includegraphics[width=\textwidth]{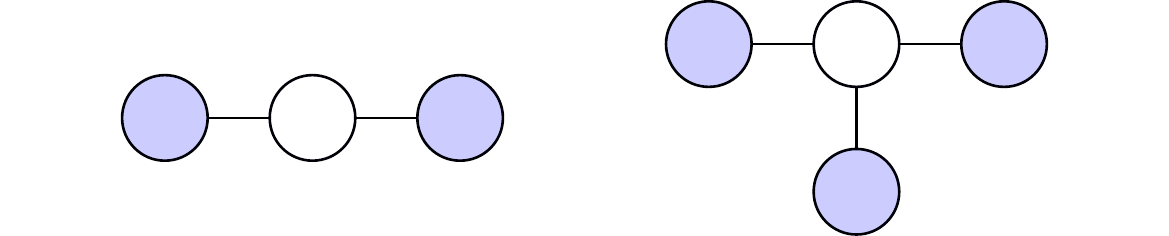}
        \caption{}
        \label{fig:beyond1wl_max}
    \end{subfigure}\hfill%
    \begin{subfigure}[t]{0.3\textwidth}
        \centering
        \includegraphics[width=\textwidth]{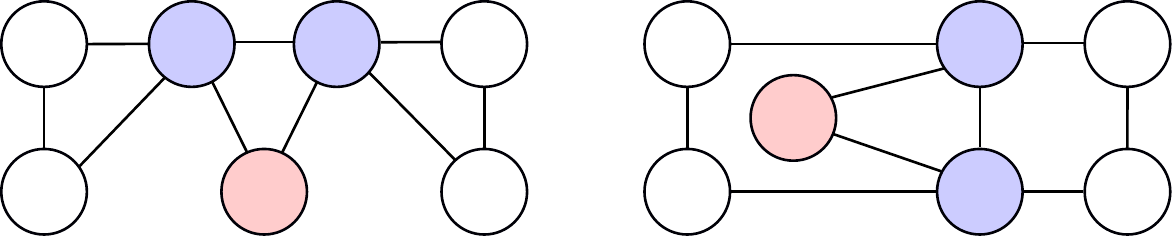}
        \caption{}
        \label{fig:beyond1wl_hard}
    \end{subfigure}\hfill%
    \caption{Graphs that cannot be distinguished by $1-$WL GNNs but by AMP.}
    \label{fig:beyond1wl}
\end{figure}

Let us now consider the scenario that delays are not random but every message takes the same time to arrive. This model is less expressive since we do not have delays as a source of randomness. Therefore, we cannot create identifiers and distinguish arbitrary graphs. Nevertheless, this variant of AMP is more powerful than the $1-$WL test. Figure~\ref{fig:beyond1wl} shows some motivational examples.

The first example (Figure~\ref{fig:beyond1wl_garg}) is one of the constructions from \citet{garg2020generalization}, where nodes have to learn if they are in the $4$ node cycles or the larger $8$ node cycle. In AMP, the node that starts can learn the cycle length and propagate the information to the rest of the graph. Furthermore, AMP can also distinguish graphs that impose constraints on the aggregation, for example, the graph in Figure~\ref{fig:beyond1wl_max} by \citet{xu2019powerful}. Since AMP receives neighborhood messages individually, the white node can react differently to each message received by a blue node. The graph in Figure~\ref{fig:beyond1wl_hard} is interesting. If we start the execution at any white or blue node, we can distinguish the two graphs easily. The top graph has a cycle of length $3$ from this starting node while the bottom graph has not. However, AMP cannot distinguish the two graphs if we start both executions at the red nodes. 

Even though AMP with constant delays is less expressive, we found, it is easier to train. With constant delays, nodes receive the same messages in the same order across different runs, which reduces noise in the gradients. On the other hand, it takes many runs with random message delays to see a representative distribution of message orderings to produce a stable gradient signal. In the subsequent experiments, we will therefore use AMP with constant delays.

\section{Experiments}
\subsection{Beyond 1-WL classification}
We experiment with AMP's ability to classify graphs that the $1-$WL test cannot distinguish. We compare on existing GNN expressiveness benchmarks for node classification (Limits1~\citep{garg2020generalization}, Limit2~\citep{garg2020generalization}, Triangles~\citep{sato2021random}, and LCC~\citep{sato2021random}) and graph classification (4-cycles~\citep{loukas20graph} and Skip-Cycles~\citep{chen2019equivalence}). Furthermore, we include two constructions by \citet{xu2019powerful} that are hard for particular aggregation and pooling functions --- MAX for \texttt{max} aggregation and MEAN for \texttt{mean} aggregation.

We compare AMP-RNN with several powerful GNN extensions from literature: PPGN~\citep{maron2019provably}, SMP~\citep{vignac2020building},\footnote{Code for SMP and PPGN from \url{https://github.com/cvignac/SMP}} DropGNN~\citep{papp2021dropgnn},\footnote{Code for GIN and DropGNN from \url{https://github.com/KarolisMart/DropGNN}} and ESAN~\citep{bevilacqua2022equivariant},\footnote{Code for ESAN from \url{https://github.com/beabevi/ESAN}} plus a simple GIN~\citep{xu2019powerful} for control. We needed to slightly modify all codebases to accept different aggregation methods. We needed to further modify ESAN to create its graph perturbations inside the model to fit into the existing framework. For training, we follow the setup by \citet{papp2021dropgnn}. GNNs have $4$ layers, for Skip-Cycles, we additionally try $9$ layers and take the better result. For AMP, we allow a total of $5n$ messages, with $n$ being the size of the graph. Figure~\ref{fig:beyond1wl_hard} shows that the starting node matters, so we execute multiple runs for AMP, one for every node with that node being the starting node. Each run computes the final embedding for the starting node. We use $16$ hidden units for Limits1, Limits2, Triangles, LCC, and 4-cycles and $32$ units for MAX, MEAN, and Skip-Cycles. Training uses the Adam~\citep{kingma2015adam} optimizer with a learning rate of $0.01$. Table~\ref{tab:symmetry} shows results on test graphs.
\begin{table*}[ht]
\centering
\resizebox{0.99\textwidth}{!}{
\begin{tabular}{@{}l*{7}{S[table-format=-3.4]}@{}}
	\toprule
{Dataset} & {GNN~\citep{xu2019powerful}} & {PPGN~\citep{maron2019provably}} & {SMP~\citep{vignac2020building}} & {DropGNN~\citep{papp2021dropgnn}} & {ESAN~\citep{bevilacqua2022equivariant}} &  {AMP (ours)}\\\midrule

{Limits1~\citep{garg2020generalization}} & \makebox{0.50 \rpm 0.00} &\makebox{0.60\rpm 0.21} & \makebox{0.95\rpm 0.16} & \makebox{\textbf{1.00\rpm 0.00}} & \makebox{\textbf{1.00\rpm0.00}} & \makebox{\textbf{1.00\rpm 0.00}}\\
{Limits2~\citep{garg2020generalization}} & \makebox{0.50\rpm 0.00} & \makebox{0.85\rpm 0.24} & \makebox{\textbf{1.00\rpm 0.00}} & \makebox{\textbf{1.00\rpm 0.00}} & \makebox{\textbf{1.00\rpm0.00}} & \makebox{\textbf{1.00\rpm 0.00}}\\
{Triangles~\citep{sato2021random}} & \makebox{0.52\rpm 0.15} & \makebox{\textbf{1.00\rpm0.02}} & \makebox{0.97\rpm 0.11} & \makebox{0.93\rpm 0.13} & \makebox{\textbf{1.00\rpm0.01}} & \makebox{\textbf{1.00\rpm0.01}} \\
{LCC~\citep{sato2021random}} & \makebox{0.38 \rpm 0.08} & \makebox{0.80\rpm0.26} & \makebox{0.95\rpm 0.17} & \makebox{\textbf{0.99\rpm 0.02}} & \makebox{\textbf{0.96\rpm0.06}} & \makebox{\textbf{0.96\rpm 0.03}} \\\midrule
{MAX~\citep{xu2019powerful}} & \makebox{0.05\rpm0.00} & \makebox{0.36\rpm0.16} & \makebox{0.74\rpm0.24} & \makebox{0.27\rpm0.07} & \makebox{0.05\rpm0.00} & \makebox{\textbf{1.00\rpm0.00}} \\
{MEAN~\citep{xu2019powerful}} & \makebox{0.28\rpm 0.31} & \makebox{0.39\rpm0.21} & \makebox{0.91\rpm0.14} & \makebox{0.58\rpm0.34} & \makebox{0.18\rpm0.08} & \makebox{\textbf{1.00\rpm0.00}} \\
{4-cycles~\citep{loukas20graph}} & \makebox{0.50\rpm 0.00} & \makebox{0.80\rpm 0.25} & \makebox{0.60\rpm0.17} & \makebox{\textbf{1.00\rpm 0.01}} & \makebox{0.50\rpm0.00} & \makebox{\textbf{1.00\rpm0.00}}\\
{Skip-Cycles~\citep{chen2019equivalence}} & \makebox{0.10\rpm 0.00} & \makebox{0.04\rpm0.07} & \makebox{0.27\rpm 0.05} & \makebox{0.82\rpm 0.28} & \makebox{0.40\rpm0.16} & \makebox{\textbf{1.00\rpm 0.00}} \\\bottomrule
\end{tabular}}
\caption{Test set accuracy on GNN expressiveness benchmarks that require beyond $1-$WL expressiveness to solve. AMP solves all benchmarks quasi-perfect, even the challenging ones that restrict the aggregation (MAX, MEAN) or require long-range propagation (Skip-Cycles).}
\label{tab:symmetry}
\end{table*}
We find that AMP performs better than all other methods and consistently solves all datasets (close to) perfectly. Many methods perform very well on the node classification tasks. On the graph classification tasks, all methods but AMP struggle. This is particularly true for the two datasets with restrictions on the aggregation. However, sometimes it is desirable not to use \texttt{sum} as aggregation function. Other aggregations, such as \texttt{max} might offer better algorithmic alignment~\citep{xu2020what, xu2021neural}. Moreover, AMP solves the challenging Skip-Cycles dataset perfectly which requires long-range information propagation.

\subsection{Long-range information propagation}
\label{sec:longrange_results}
In this section, we investigate the long-range information propagation of AMP. We experiment with a simplified version of finding shortest paths in graphs. Finding shortest paths is interesting for long-range information propagation since it requires reading the entire graph in the worst case and has been used previous works~\citep{tang2020towards, velickovic2020neural, xu2021neural}. We simplify the shortest path setting: instead of regressing the exact distance, we classify if the shortest path is even or odd. We do this to abstract from the need to do accurate computation of distances. Neural networks struggle with these arithmetic computations in general~\citep{faber2020neural, heim2020neural, madsen2020neural, trask2018neuralalu}. We compare AMP with several synchronous GNNs (IterGNNs~\citep{tang2020towards},\footnote{Code from IterGNN from \url{https://github.com/haotang1995/IterGNN}} Universal Transformers~\citep{dehghani2018universal}, and NEG (Neural Execution of Graph Algorithms~\citep{velickovic2020neural}). For GNNs, we mark the starting node by giving it distinct features, for AMP, we sent this node the initial message.

We train on $25$ randomly create graphs with $10$ trees. Graphs are based on a spanning tree to which we add $\frac{n}{5}$ extra edges. Training runs for $1000$ iterations and uses the Adam optimizer with a learning rate of $0.01$. The hidden size dimension is $30$. While we train on graphs with size $10$, we follow previous work~\citep{tang2020towards, velickovic2020neural, xu2021neural} to test the ability to extrapolate the solution to larger graphs. We test with graphs of sizes $10$, $15, 25, 50, 100, 250, 500, 1000$. Larger graphs are also more challenging since the shortest paths grow in length. We report the classification accuracy per size in Table~\ref{tab:nonlocal}.

We can see that all AMP version perform better than the synchronous baseline GNNs. AMP-Iter performs especially well. We hypothesize that this is because the asynchronous model aligns better with the given task. Every node only needs to act when there might be relevant information. On the other hand, nodes in the synchronous GNNs need to stay ready over many rounds until the information from the starting node finally reaches them. \citet{xu2020what} and \citet{xu2021neural} have shown that better algorithmic alignment between the neural architecture and the task helps the model to learn with fewer samples and extrapolate better.

\begin{table*}[ht]
\centering
\resizebox{0.99\textwidth}{!}{
\begin{tabular}{@{}l*{8}{S[table-format=-3.4]}@{}}
\toprule
{Model} & {10} & {25} & {50} & {100} & {250} & {500} & {1000} & {2500}\\\midrule
{NEG~\citep{velickovic2020neural}} & \makebox{0.67\rpm0.12} & \makebox{0.54\rpm0.08} & \makebox{0.51\rpm0.03} & \makebox{0.51\rpm0.01} & \makebox{0.49\rpm0.01} & \makebox{0.50\rpm0.01} & \makebox{0.50\rpm0.00} & \makebox{0.50\rpm0.00}\\
{Universal\citep{dehghani2018universal}} & \makebox{0.91\rpm0.05} & \makebox{0.74\rpm0.09} & \makebox{0.65\rpm0.11} & \makebox{0.61\rpm0.13} & \makebox{0.57\rpm0.14} & \makebox{0.56\rpm0.14} & \makebox{0.55\rpm0.14} & \makebox{0.55\rpm0.14}\\
{IterGNN~\citep{tang2020towards}} & \makebox{0.92\rpm0.10} & \makebox{0.82\rpm0.15} & \makebox{0.73\rpm0.13} & \makebox{0.65\rpm0.09} & \makebox{0.58\rpm0.05} & \makebox{0.55\rpm0.03} & \makebox{0.52\rpm0.01} & \makebox{0.51\rpm0.00} \\\midrule
{AMP-RNN} & \makebox{0.96\rpm0.02} & \makebox{0.85\rpm0.05} & \makebox{0.74\rpm0.08} & \makebox{0.69\rpm0.07} & \makebox{0.61\rpm0.09} & \makebox{0.58\rpm0.09} & \makebox{0.56\rpm0.05} & \makebox{0.54\rpm0.04} \\
{AMP-GRU} & \makebox{0.97\rpm0.01} & \makebox{0.89\rpm0.03} & \makebox{0.80\rpm0.05} & \makebox{0.73\rpm0.10} & \makebox{0.66\rpm0.07} & \makebox{0.63\rpm0.08} & \makebox{0.60\rpm0.05} & \makebox{0.56\rpm0.04} \\
{AMP-LSTM} & \makebox{0.97\rpm0.01} & \makebox{0.87\rpm0.03} & \makebox{0.77\rpm0.06} & \makebox{0.69\rpm0.10} & \makebox{0.61\rpm0.09} & \makebox{0.58\rpm0.10} & \makebox{0.56\rpm0.06} & \makebox{0.54\rpm0.05} \\\midrule
{AMP-ACT} & \makebox{\textbf{1.00\rpm0.00}} & \makebox{0.98\rpm0.02} & \makebox{0.96\rpm0.04} & \makebox{0.95\rpm0.05} & \makebox{0.93\rpm0.07} & \makebox{0.91\rpm0.10} & \makebox{0.91\rpm0.11} & \makebox{0.90\rpm0.12} \\
{AMP-Iter} & \makebox{\textbf{1.00\rpm0.00}} & \makebox{\textbf{1.00\rpm0.00}} & \makebox{\textbf{1.00\rpm0.00}} & \makebox{\textbf{1.00\rpm0.00}} & \makebox{\textbf{1.00\rpm0.00}} & \makebox{\textbf{0.99\rpm0.00}} & \makebox{\textbf{0.99\rpm0.00}} & \makebox{\textbf{0.99\rpm0.00}}\\\bottomrule
\end{tabular}}
\caption{Accuracy for predicting the parity of shortest paths to a starting node. The table head contains the number of nodes in the test graph while training is always on $10$ nodes. AMP-Iter learns to extrapolate almost perfectly. Also other AMP variants extrapolate better than the GNN baselines.}
\label{tab:nonlocal}
\end{table*}%

\begin{figure}[h]
    \begin{subfigure}[t]{0.48\textwidth}
        \centering
        \includegraphics[width=\textwidth]{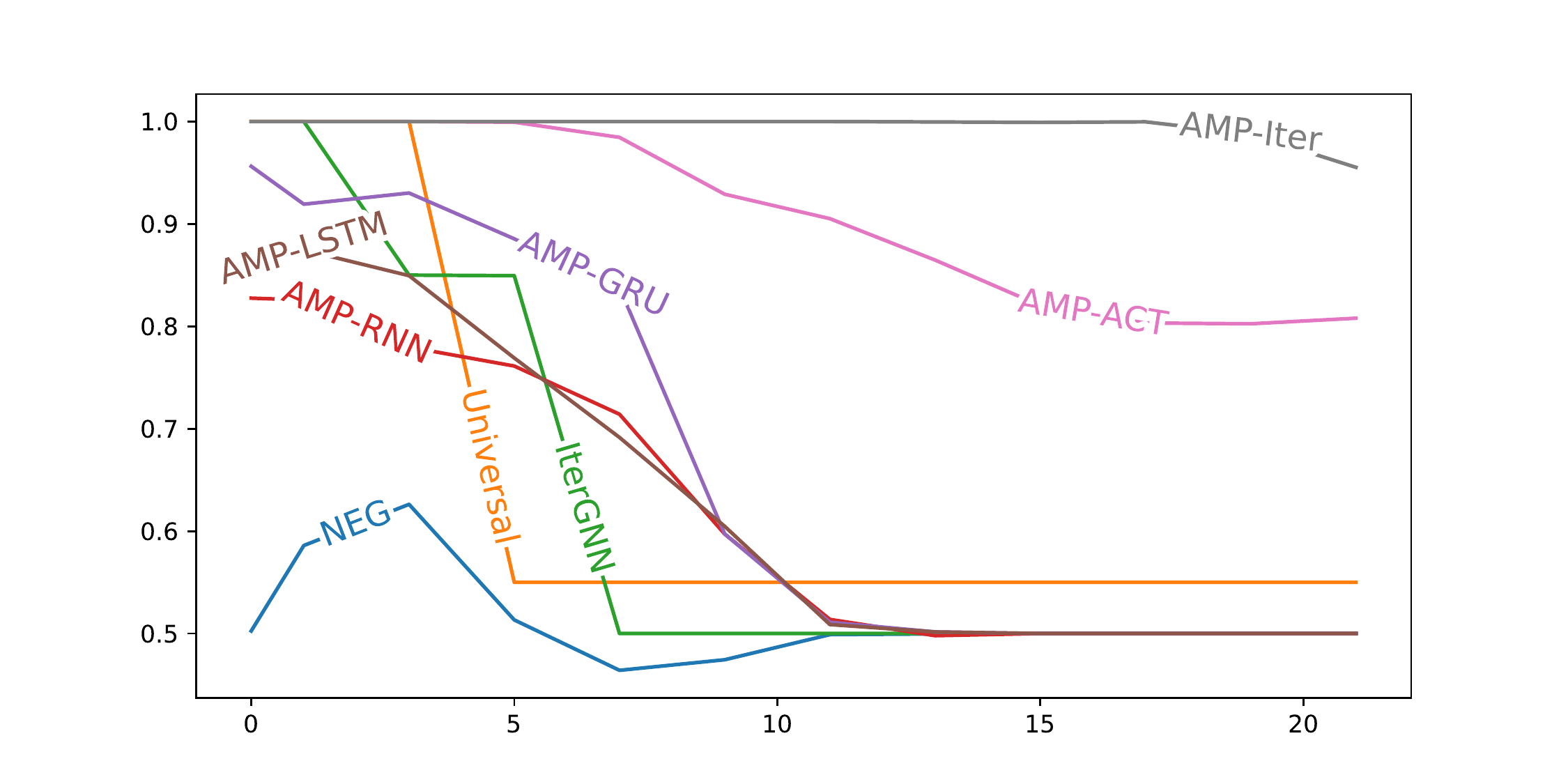}
        \caption{Accuracy for the shortest path parity ($y-$axis). We break accuracy down by distance to the starting node ($x-$axis). The stronger accuracy declines for increasing $x$, the more a method is exposed to underreaching.}
        \label{fig:underreaching}
    \end{subfigure}\hfill%
    \begin{subfigure}[t]{0.48\textwidth}
        \centering
        \includegraphics[width=\textwidth]{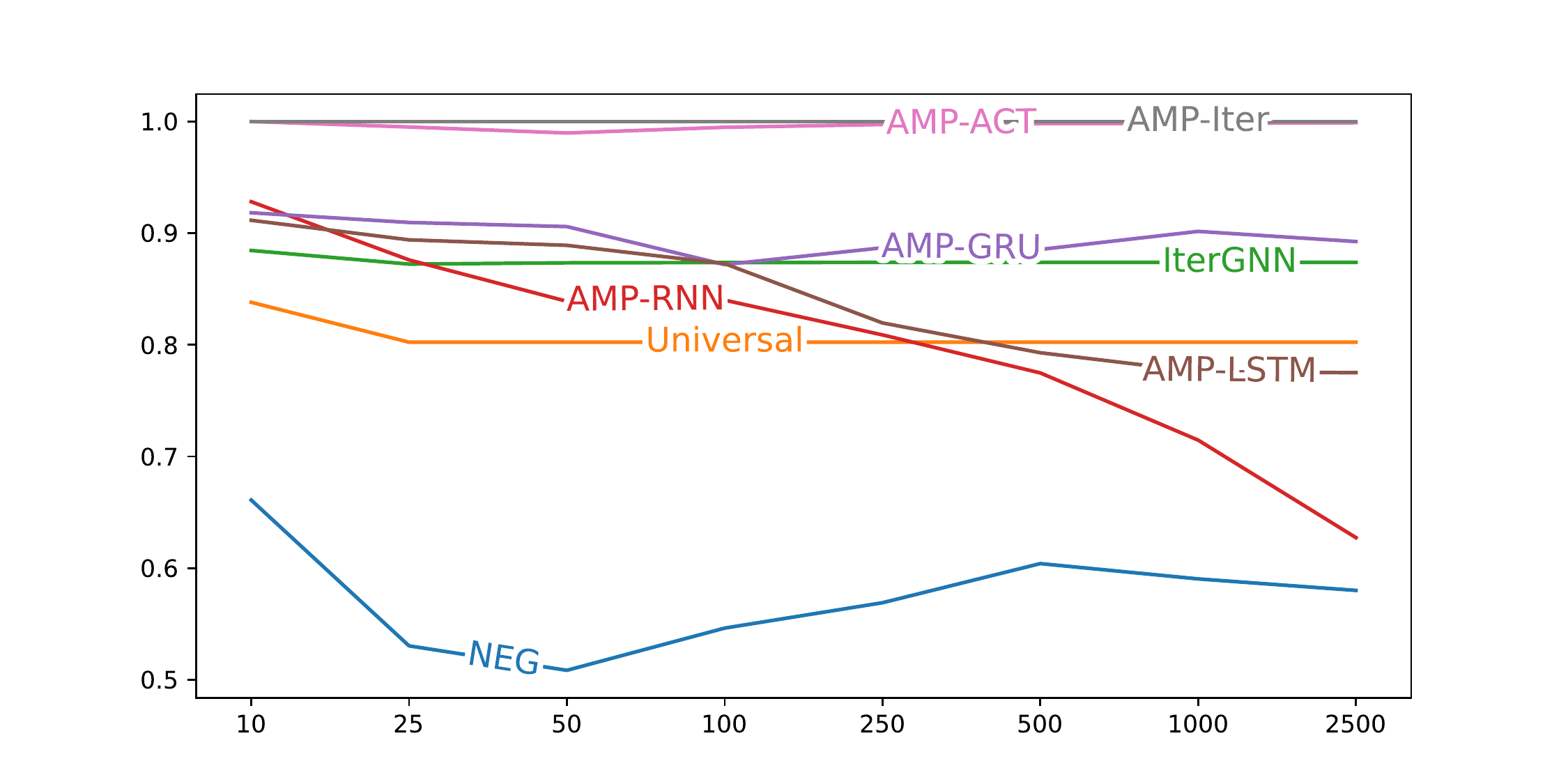}
        \caption{Accuracy for the shortest path parity ($y-$axis) but only for nodes with a distance the model saw in the training set for different graph sizes ($x-$axis). A declining accuracy suggests oversmoothing.}
        \label{fig:oversmoothing}
    \end{subfigure}\hfill%
\end{figure}

We now investigate if we can understand the accuracy losses of Table~\ref{tab:nonlocal} in terms of underreaching and oversmoothing. To estimate underreaching, we break down the accuracy by distance from the starting node. If accuracy decreases with increasing distance it suggests exposure to underreaching. Practically, since the label flips in every step we merge an odd and an even pair into one bucket. For example the scores for distances $1$ and $2$ are combined. Figure~\ref{fig:underreaching} shows the results broken down by distance. We can see that most all architectures except AMP-Iter and AMP-ACT cannot extrapolate to much larger distances than the training set. This contributes largely to the decrease in accuracy since the larger graphs become, the more nodes have large distances to the starting node.

To estimate oversmoothing, we restrict the accuracy measurement towards nodes whose distance to the starting node also is in the training set. If the accuracy relatively decreases, the presence of further away nodes impacts close-by nodes, which indicates oversmoothing. Figure~\ref{fig:oversmoothing} shows the results for this analysis. In principle, all models combat overmsoothing well except AMP-RNN.

\begin{figure}[h]
    \begin{subfigure}[t]{0.48\textwidth}
        \centering
        \includegraphics[width=\textwidth]{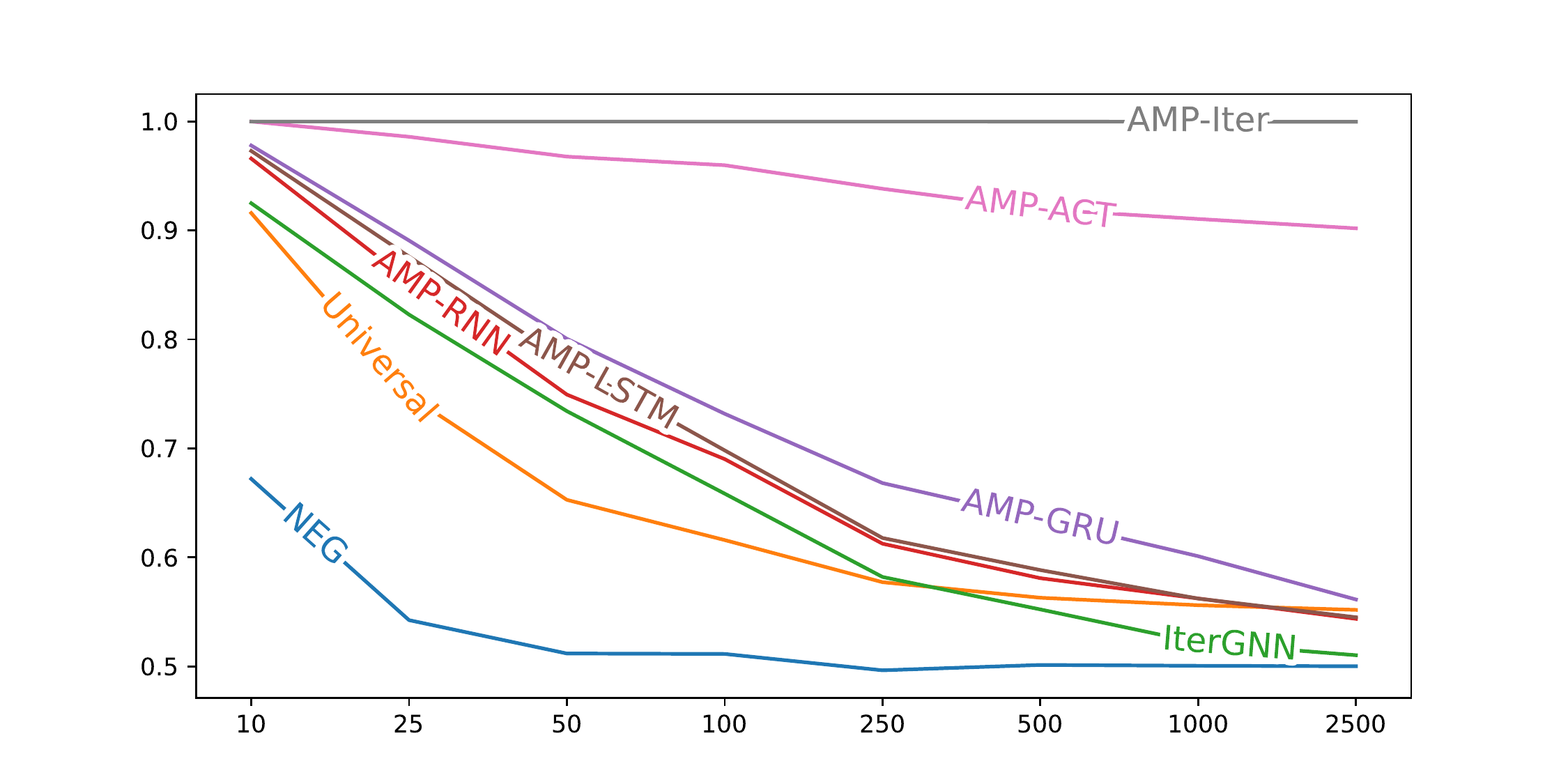}
        \caption{}
        \label{fig:oversquashing1}
    \end{subfigure}\hfill%
    \begin{subfigure}[t]{0.48\textwidth}
        \centering
        \includegraphics[width=\textwidth]{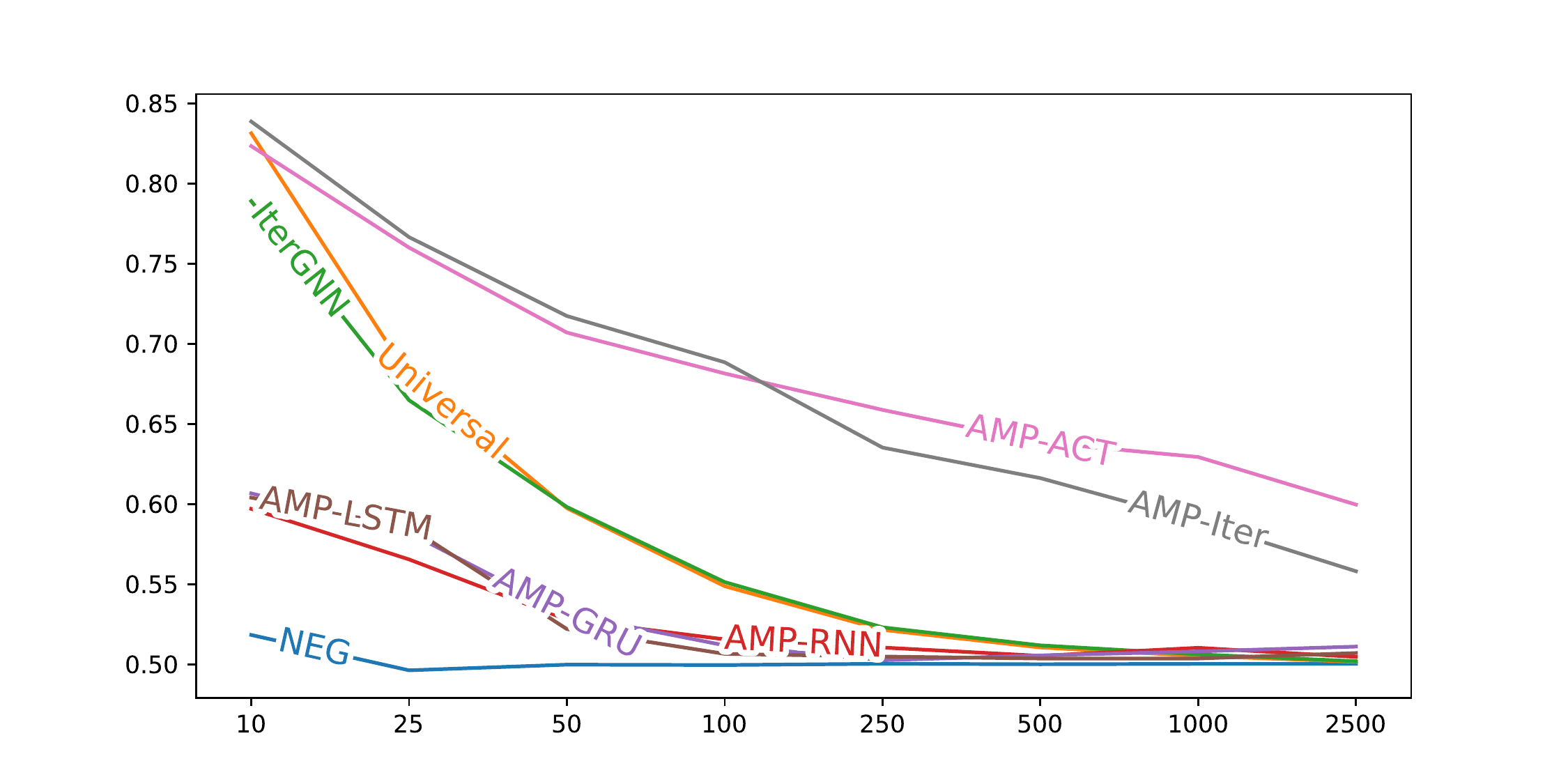}
        \caption{}
        \label{fig:oversquashing2}
    \end{subfigure}\hfill%
    \caption{Accuracy for the shortest path parity task ($y-axis$) for solving (a) one or (b) three tasks at the same time for different graph sizes ($x-$axis). The more accuracy drops between the left to the right figure, the more a method is exposed to oversquashing.}
    \label{fig:oversquashing}
\end{figure}

Third we estimate the expose of different models to oversquashing. We compare the performance of the model when it has to solve the shortest path task to when it has to solve three shortest path tasks at the same time on the same graph. This triples the amount of information nodes need to exchange. Figure~\ref{fig:oversquashing} shows the accuracy for the single problem, Figure~\ref{fig:oversquashing2} the accuracy for the triple problem.

We see that all methods deteriorate in performance, which is expected since the learning task is harder. But we can see that different methods deteriorate in different speed. Again AMP-Iter and AMP-ACT perform best. Since nodes in AMP receive information for one problem at a time, they are less exposed to oversquashing. Learning to terminate also helps nodes to keep the information they learnt. Recurrent AMP variants perform worse than AMP-Iter and AMP-ACT and even worse than IterGNN and Universal. This suggests that learning to terminate is indeed a useful property.

\subsection{Graph Classification}
Last, we train AMP on several graph classification benchmarks. Our AMP implementation runs on a single CPU since the sequential nature limits GPU potential and python's Global-Interpreter-Lock\footnote{\url{https://wiki.python.org/moin/GlobalInterpreterLock}} prevents multithreading. This makes training larger datastes prohibitively slow for now (for example, PROTEINS takes around $1-2$ day to train). Therefore we limit the comparison to the smaller datasets MUTAG, PTC, PROTEINS, IMDB-B, IMDB-M~\citep{yanardag2015deep}. Further, we do a single pass in each training epoch instead of $50$ batches of $50$ graphs. We run AMP-RNN that does a short run from every node. We do a small grid search over the number of messages ($15$ or $25$) per run, the size of the message embeddings ($10$ or half the node embedding size) and whether AMP-RNN uses skip connections to previous states. The remaining setup is taken from \citet{xu2019powerful}: We run on $10$ different splits and report the accuracy of the best performing epoch. We use hidden node states of $64$ for social datasets and $16$ or $32$ for biological datasets. We further compute results for GraphSAGE~\citep{hamilton2017inductive}, GCN~\citep{kipf2017semi}, and GAT~\citep{velivckovic2018graph} and take results for more GNNs. Table~\ref{tab:graph_classification} shows all results.

Even with little investigation into suitable AMP architectures and hyperparameter tuning, AMP achieves comparable results to existing GNN works but not quite state of the art. We believe that further improvements in AMP have the potential to reach a competitive performance. What surprised us are the results of GCN that clearly outperform all other methods on $3$ out of $5$ datasets.

\begin{table*}[ht]
\centering
\resizebox{0.75\textwidth}{!}{
\begin{tabular}{@{}l*{6}{S[table-format=-3.4]}@{}}
\toprule
{Model} & {MUTAG} & {PTC} & {PROTEINS} & {IMDB-B} & {IMDB-M} & \\
\midrule  
{PatchySan~\citep{niepert2016learning}} & \makebox{\textbf{92.6 \rpm 4.2}} & \makebox{62.3 \rpm 5.7} &  \makebox{75.9 \rpm 2.8} & \makebox{71.0 \rpm 2.2} & \makebox{45.2 \rpm 2.8}\\
{DGCNN~\citep{zhang2018end}} & \makebox{85.8 \rpm 1.7} & \makebox{58.6 \rpm 2.5} &  \makebox{75.5 \rpm 0.9} & \makebox{70.0 \rpm 0.9} & \makebox{47.8 \rpm 0.9}\\
{GraphSAGE~\citep{hamilton2017inductive}} & \makebox{90.4\rpm7.8} & \makebox{63.7\rpm 9.7} &  \makebox{75.6\rpm 5.5} & \makebox{76.0\rpm 3.3} & \makebox{51.9\rpm 4.9}\\
{GCN~\citep{kipf2017semi}} & \makebox{88.9 \rpm 7.6} & \makebox{\textbf{79.1 \rpm 11.4}} &  \makebox{76.9 \rpm 4.8} & \makebox{\textbf{83.4 \rpm 4.9}} & \makebox{\textbf{57.5\rpm 2.6}}\\
{GAT~\citep{velivckovic2018graph}} & \makebox{85.1 \rpm 9.3} & \makebox{64.5\rpm 7.0} &  \makebox{75.4\rpm3.8} & \makebox{74.9\rpm 3.8} & \makebox{52.0\rpm3.0}\\
{GIN~\citep{xu2019powerful}} & \makebox{89.4 \rpm 5.6} & \makebox{66.6 \rpm 6.9} &  \makebox{76.2 \rpm 2.6} & \makebox{75.1 \rpm 5.1} & \makebox{52.3 \rpm 2.8}\\\midrule
{AMP-RNN (ours)} & \makebox{90.4\rpm4.1} & \makebox{63.7 \rpm 9.1} & \makebox{76.7\rpm 7.1} & \makebox{74.6\rpm 3.6} & \makebox{52.1\rpm 3.6}\\\midrule
{1-2-3 GNN~\citep{morris2019weisfeiler}} & \makebox{86.1} & \makebox{60.9} &  \makebox{75.5} & \makebox{74.2} & \makebox{49.5}\\
{DropGNN~\citep{papp2021dropgnn}} & \makebox{90.4 \rpm 7.0} & \makebox{66.0\rpm 9.8} &  \makebox{76.3 \rpm 6.1} & \makebox{75.7 \rpm 4.2} & \makebox{51.4 \rpm 2.8}\\
{PPGN~\citep{maron2019provably}*} & \makebox{90.6 \rpm 8.7} & \makebox{66.2 \rpm 6.5} &  \makebox{77.2 \rpm 4.7} & \makebox{73 \rpm 5.8} & \makebox{50.5 \rpm 3.6}\\
{ESAN~\citep{bevilacqua2022equivariant}*} & \makebox{92.0\rpm5.0} & \makebox{69.2\rpm 6.5} & \makebox{\textbf{77.3 \rpm 3.8}} & \makebox{77.1\rpm 2.6} & \makebox{53.7\rpm 2.1}\\\bottomrule
\end{tabular}}
\caption{Graph classification accuracy (\%). AMP produces results that are comparable to simple and new more expressive GNN variants. *We report the result achieved by the best model version model.}  
\label{tab:graph_classification}
\end{table*}

\section{Conclusion and Impact Statement}
In this paper, we presented a new paradigm for learning neural networks on graphs. In GNNs, which are inspired by synchronous distributed algorithms, every node sends a message to every neighbor in every round. Nodes aggregate all messages before parsing them, which leads to problems such as oversmoothing, underreaching, oversquashing, and being restricted in expressiveness by the $1-$WL test. We present a new framework, AMP, that is inspired by the asynchronous communication model. We theoretically investigate its expressiveness and empirically evaluate its ability to distinguish hard graphs, propagate information over long distances, and classify common benchmarks. AMP might also be more interpretable if we investigate when nodes react or ignore a message.

\textbf{Impact Statement.} We believe that AMP is a promising paradigm. GNNs have received a lot of attention over the years. New GNN variants have improved the original suggestions considerably. We believe that AMP deserves the same attention, and will improve accordingly. But before we can apply AMP to most real-world problems, we need improve tooling to allow parallelism and larger graphs.

\bibliography{paper}
\bibliographystyle{abbrvnat}

\clearpage

\appendix

\clearpage
\section{Weights}
\label{sec:simulation_weights}
\begin{figure}[h]
    \centering
    \includegraphics[width=0.8\textwidth]{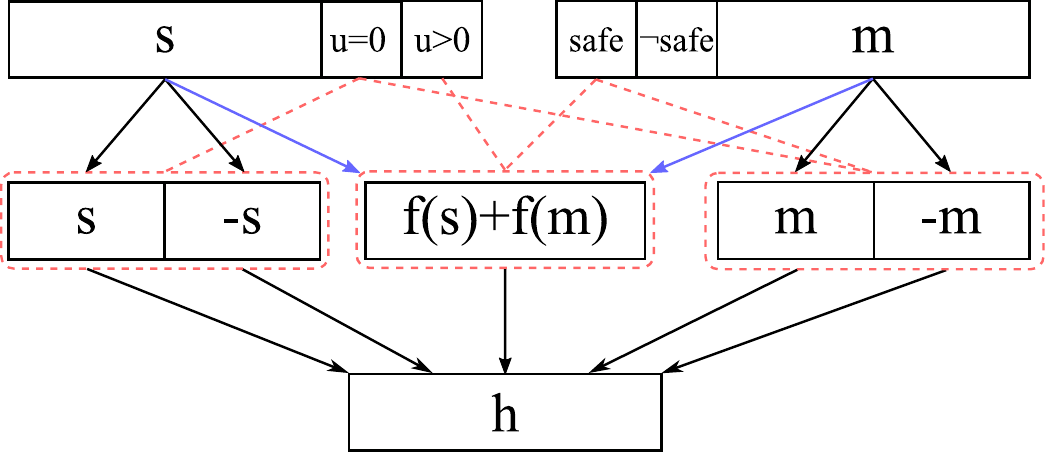}
    \caption{Weight construction to create a $3-$way \texttt{if}. Solid black lines are $+1$ or $-1$ and the solid blue lines are the weights of sGIN. Dotted red lines correspond to large negative weights, if the binary value belonging to them is $1$, they override any content to a negative value that gets zeroed by the ReLu activation.}
    \label{fig:ifgate}
\end{figure}

This section is a follow-up to the proof for Theorem~\ref{theorem:simulation}. We will show that a $3-$ layer MLP with a ReLu activation can exactly model the transition function in Table~\ref{tab:f_table_simulation}. This is because the transition function is an ordered sequence of \texttt{if} statements. Our construction is as follows: The first layer adds the complements for every bit. For example, if we have the \texttt{pulse} bit, the layer adds $\neg \texttt{pulse}$. Furthermore, we add two bits for counts such as $w$: $w=0$ and $w > 0$. Let $f$ be the linear layer from sGIN and we want to compute the following.
\begin{equation*}
    f = \begin{cases}
    s & \text{ if } \texttt{safe}=1\\
    ReLu(f(s+m)) & \text{ if } u=0\\
    s+m & \text{else}
    \end{cases}
\end{equation*}
The function is a subset of the function in Table~\ref{tab:f_table_simulation}. The first case preserves the state, the second case updates the state with the same function as sGIN and the third state adds the message to the state. Having bit complements and count bits from the first layer, we create the last $2$ layers as in Figure~\ref{fig:ifgate}. The middle row shows the three components that we have for the three cases. The left component stores $s$, the right component stores $m$, and the middle component simulates sGIN. The key motif in this construction is attaching large negative weights from bits to the components these bits disable. If a bit (or its complement) is set, the component will become negative and set to $0$ by ReLu. For example, if $safe=1$, both the middle and right components are zeroed, so that $h=s$.

\clearpage
\section{Proof for Lemma~\ref{lemma:identifiers}}
\label{sec:proof_identifiers}
We show that AMP with random delays can crate unique identifiers for every node in a star graph with central node $v$ and $k$ outer nodes. If delays are randomly from the uniform distribution [0, 1], ID creation takes $O(k\cdot 6^k)$ time.

The idea in the assignment algorithm is that the outer nodes compete for IDs in a first come first served fashion. Nodes can get lucky with the delays to beat other nodes. To allow direct competition, we add edges between every outer node.\footnote{These edges are not necessary since the central node $v$ can also propagate messages between outer nodes, but they make the idea more concise and the algorithm faster} These competitions are not always successful therefore every node has a state \texttt{try} to keep track of the current attempt and the \texttt{ID} they eventually receive. The central node $v$ is always in state \texttt{assigning} until every outer node received an ID. Then $v$ switches to \texttt{having}. The central node additionally keeps track of the number of nodes $w$ that it is waiting for a reply for the current try and the number of nodes $x$ that do not yet have an ID. The outer nodes that do not have an ID use the states \texttt{taking} to try to get an ID or \texttt{yielding} to acknowledge another node was faster to claim. The central keeps track of the number $c$ that are trying to claim the ID in the current try. We create one bit each for every of the four possible states which gives a state of $(\texttt{try}, \texttt{ID}, c, w, x \texttt{assigning}, \texttt{having}, \texttt{taking}, \texttt{yielding})$ for all nodes, initially every node is in state $(\texttt{-1, 0, 0, 0, 0, \texttt{yielding}})$.

Nodes communicate via one of four types of message types that are represented via bits. A message can \texttt{offer} and ID, \texttt{confirm} an ID, \texttt{claim} an ID, or \texttt{surrender} an ID. Every message also contains the candidate ID (\texttt{CID}) and attempt in question. Again, the very first message requires special handling is marked by a special \texttt{origin} bit. This creates messages of the type $(\texttt{CID}, \texttt{attempt}, \texttt{offer}, \texttt{confirm}, \texttt{claim}, \texttt{surrender}, \texttt{origin})$.

Tables~\ref{tab:f_table_id_center} and \ref{tab:f_table_id_outer} show the transition functions for a node in state $(\texttt{try}, \texttt{ID}, c, w, x, \texttt{state})$ and a message $(\texttt{CID}, \texttt{attempt}, \texttt{type})$. For conciseness, we separated the actions of the central versus outer nodes but the two tables are actually one function. Every node can identify if it is central, then it would be in state \texttt{assigning}, otherwise the node is an outer node.

\begin{table*}[ht]
\centering
\resizebox{0.99\textwidth}{!}{
\begin{tabular}{@{}l*{10}{S[table-format=-3.4]}@{}}
\toprule
{Condition} & {\texttt{state}} & {\texttt{try}} & {\texttt{ID}} & {c} & {w} & {x} & {\texttt{CID}} & {\texttt{attempt}} & {\texttt{type}}\\\midrule
{origin=1} & {\texttt{assigning}} & {$0$} & {$0$} & {$0$} & {$D-1$} & {$D-1$} & {$1$} & {$0$} & {\texttt{offer}}\\
{\texttt{try} $\neq$ \texttt{attempt}} & {\texttt{state}} & {\texttt{try}} & {\texttt{0}} & {c} & {w} & {x} & {$\bot$} & {$\bot$} & {$\bot$} \\
{\texttt{claim} $\land ~c=0$} & {\texttt{assigning}} & {\texttt{try}} & {$0$} & {$1$} & {$w-1$} & {$x$} & {$\bot$} & {$\bot$} & {$\bot$}\\
{\texttt{claim} $\land ~c>1$} & {\texttt{assigning}} & {\texttt{try}+1} & {$0$} & {$0$} & {$x-1$} & {$x$} & {\texttt{CID}} & {\texttt{try}+1} & {\texttt{offer}}\\
{\texttt{surrender}} & {\texttt{assigning}} & {\texttt{try}} & {$0$} & {$c$} & {$w-1$} & {$x$} & {$\bot$} & {$\bot$} & {$\bot$}\\
{$w=0 \land x>0$} & {\texttt{assigning}}  & {\texttt{try}+1} & {$0$} & {$0$} & {$x-2$} & {$x-1$} & {\texttt{CID}+1} & {\texttt{try}+1} & {\texttt{offer}}\\
{$x=0$} & {\texttt{having}} & {$0$} & {$0$} & {$0$} & {$0$} & {$0$} & {$0$} & {$0$} & {\texttt{confirm}}\\\bottomrule
\end{tabular}}
\caption{Update function $f$ used the center node of a star in AMP to create unique identifiers for every node. The center node always assumes and keeps ID $0$.}
\label{tab:f_table_id_center}

\resizebox{0.99\textwidth}{!}{
\begin{tabular}{@{}l*{10}{S[table-format=-3.4]}@{}}
\toprule
{Condition} & {state} & {try} & {ID} & {c} & {w} & {x} & {CID} & {attempt} & {type}\\\midrule
{$\texttt{offer} \land \texttt{attempt} > \texttt{try}$} & {\texttt{taking}} & {\texttt{attempt}} & {\texttt{CID}} & {$0$} & {$0$} & {$0$} & {\texttt{CID}} & {\texttt{attempt}} & {\texttt{claim}} \\
{\texttt{taking} $\land \texttt{attempt} > \texttt{try}$} & {\texttt{yielding}} & {\texttt{attempt}} & {$0$} & {$0$} & {$0$} & {$0$} & {\texttt{CID}} & {\texttt{attempt}} & {\texttt{surrender}}\\
{\texttt{taking} $\land \texttt{offer} \land \texttt{ID}\neq \texttt{CID}$} & {\texttt{having}} & {$0$} & {\texttt{ID}} & {$0$} & {$0$} & {$0$} & {$\bot$} & {$\bot$} & $\bot$\\
{\texttt{taking} $\land \texttt{confirm}$} & {\texttt{having}} & {$0$} & {\texttt{ID}} & {$0$} & {$0$} & {$0$} & {$\bot$} & {$\bot$} & $\bot$\\\bottomrule
\end{tabular}}
\caption{Update function $f$ used the outer nodes of a star in AMP to create unique identifiers for every node.}
\label{tab:f_table_id_outer}
\end{table*}
\label{tab:f_table_id}

\begin{proof}
We need to show that this function assigns every node a unique ID. Center node $v$ assumes and keeps the ID $0$ and only proposes IDs of $1$ and greater to other nodes, so $v$'s ID is unique. We need to show that (i) no two outer nodes receive the same ID, and (ii) every outer node receives an ID.

(i) Before a node $n$ receives an ID from $v$ it has to claim the ID and make this claim known to $v$ and all other nodes. If all other nodes surrender to $n$'s claim they send $v$ a \texttt{surrender} message. Only if all other nodes without ID surrender, $v$ confirms the ID to $n$ (either implicitly by offering the next ID or explicitly by sending a \texttt{confirm} message.

(ii) Node $v$ stops giving out IDs when its internal counter $x$ reaches $0$. This counter decreases when $v$ confirms an ID, therefore $v$ confirms one ID for every neighbor. No neighbor claims two IDs so every neighbor receives an ID.

This function is also a prioritized sequence of if statements. We can use the same construction idea from Appendix~\ref{sec:simulation_weights} to exactly model this function with a $3-$layer MLP.

To understand the time complexity let us look at the probability for one node $n$ to \texttt{surrender} to another node $m$. The delays of the \texttt{offer} message from $v$ to $m$ plus the delay of the \texttt{claim} message from $m$ to $n$ need to be smaller than the delay for the \texttt{offer} message from $v$ to $n$. For uniformly random delays from [0, 1], this probability is $\frac{1}{6}$. For a node to receive an ID, all other $k-1$ neighbors without an ID need to \texttt{surrender}, which has probability $(\frac{1}{6})^{(k-1)}$.

\end{proof}

\clearpage
\section{Tabular results for Section~\ref{sec:longrange_results}}
In this section, we attach the data underlying the figures in Section~\ref{sec:longrange_results} plus the standard deviations, that do not fit the figures in that section.

\begin{table*}[ht]
\centering
\resizebox{0.99\textwidth}{!}{
\begin{tabular}{@{}l*{8}{S[table-format=-3.4]}@{}}
\toprule
{Distance} & {NEG} & {Universal} & {IterGNN} & {AMP-RNN} & {AMP-GRU} & {AMP-LSTM} & {AMP-ACT} & {AMP-Iter}\\\midrule
0-1 & \makebox{0.58\rpm0.06} & \makebox{\textbf{1.00\rpm0.00}} & \makebox{\textbf{1.00\rpm0.00}} & \makebox{0.82\rpm0.07} & \makebox{0.91\rpm0.07} & \makebox{0.87\rpm0.07} & \makebox{\textbf{1.00\rpm0.00}} & \makebox{\textbf{1.00\rpm0.00}}\\
2-3 & \makebox{0.62\rpm0.07} & \makebox{\textbf{1.00\rpm0.00}} & \makebox{0.85\rpm0.11} & \makebox{0.77\rpm0.07} & \makebox{0.93\rpm0.06} & \makebox{0.84\rpm0.08} & \makebox{\textbf{1.00\rpm0.00}} & \makebox{\textbf{1.00\rpm0.00}}\\
4-5 & \makebox{0.51\rpm0.02} & \makebox{0.55\rpm0.07} & \makebox{0.84\rpm0.11} & \makebox{0.76\rpm0.09} & \makebox{0.88\rpm0.10} & \makebox{0.76\rpm0.11} & \makebox{\textbf{0.99\rpm0.00}} & \makebox{\textbf{1.00\rpm0.00}}\\
6-7 & \makebox{0.46\rpm0.04} & \makebox{0.55\rpm0.07} & \makebox{0.50\rpm0.00} & \makebox{0.71\rpm0.11} & \makebox{0.83\rpm0.07} & \makebox{0.69\rpm0.15} & \makebox{\textbf{0.98\rpm0.01}} & \makebox{\textbf{1.00\rpm0.00}}\\
8-9 & \makebox{0.47\rpm0.03} & \makebox{0.55\rpm0.07} & \makebox{0.50\rpm0.00} & \makebox{0.59\rpm0.06} & \makebox{0.59\rpm0.05} & \makebox{0.60\rpm0.07} & \makebox{0.92\rpm0.07} & \makebox{\textbf{1.00\rpm0.00}}\\
10-11 & \makebox{0.49\rpm0.00} & \makebox{0.55\rpm0.07} & \makebox{0.50\rpm0.00} & \makebox{0.51\rpm0.00} & \makebox{0.51\rpm0.00} & \makebox{0.50\rpm0.01} & \makebox{0.90\rpm0.07} & \makebox{\textbf{1.00\rpm0.00}}\\
12-13 & \makebox{0.49\rpm0.00} & \makebox{0.55\rpm0.07} & \makebox{0.50\rpm0.00} & \makebox{0.49\rpm0.00} & \makebox{0.50\rpm0.00} & \makebox{0.50\rpm0.00} & \makebox{0.86\rpm0.07} & \makebox{\textbf{0.99\rpm0.00}}\\
14-15 & \makebox{0.50\rpm0.00} & \makebox{0.55\rpm0.07} & \makebox{0.50\rpm0.00} & \makebox{0.50\rpm0.00} & \makebox{0.50\rpm0.00} & \makebox{0.50\rpm0.00} & \makebox{0.82\rpm0.09} & \makebox{\textbf{0.99\rpm0.00}}\\
16-17 & \makebox{0.50\rpm0.00} & \makebox{0.55\rpm0.07} & \makebox{0.50\rpm0.00} & \makebox{0.50\rpm0.00} & \makebox{0.50\rpm0.00} & \makebox{0.50\rpm0.00} & \makebox{0.80\rpm0.09} & \makebox{\textbf{0.99\rpm0.00}}\\
18-19 & \makebox{0.50\rpm0.00} & \makebox{0.55\rpm0.07} & \makebox{0.50\rpm0.00} & \makebox{0.50\rpm0.00} & \makebox{0.50\rpm0.00} & \makebox{0.50\rpm0.00} & \makebox{0.80\rpm0.09} & \makebox{\textbf{0.98\rpm0.01}}\\
20-21 & \makebox{0.50\rpm0.00} & \makebox{0.55\rpm0.07} & \makebox{0.50\rpm0.00} & \makebox{0.50\rpm0.00} & \makebox{0.50\rpm0.00} & \makebox{0.50\rpm0.00} & \makebox{0.80\rpm0.10} & \makebox{\textbf{0.95\rpm0.06}}\\\bottomrule
\end{tabular}}
\caption{Investigation of the underreaching effect. Table columns show the accuracy per metric for nodes with increasing distance to the starting node. If a loss in accuracy occurs, the model has trouble transmitting the information to larger distances. AMP-based approaches are better equipped to transmit information to long distances, in particular AMP-Iter.}
\label{tab:nonlocal_underreaching}
\end{table*}%

\begin{table*}[ht]
\centering
\resizebox{0.99\textwidth}{!}{
\begin{tabular}{@{}l*{8}{S[table-format=-3.4]}@{}}
\toprule
{Model} & {10} & {25} & {50} & {100} & {250} & {500} & {1000} & {2500}\\\midrule
{NEG~\citep{velickovic2020neural}} & \makebox{0.66\rpm0.11} & \makebox{0.53\rpm0.11} & \makebox{0.50\rpm0.07} & \makebox{0.54\rpm0.07} & \makebox{0.56\rpm0.11} & \makebox{0.60\rpm0.13} & \makebox{0.59\rpm0.13} & \makebox{0.58\rpm0.14}\\
{Universal\citep{dehghani2018universal}} & \makebox{0.83\rpm0.06} & \makebox{0.80\rpm0.08} & \makebox{0.80\rpm0.08} & \makebox{0.80\rpm0.08} & \makebox{0.80\rpm0.08} & \makebox{0.80\rpm0.08} & \makebox{0.80\rpm0.08} & \makebox{0.80\rpm0.08}\\
{IterGNN~\citep{tang2020towards}} & \makebox{0.88\rpm0.14} & \makebox{0.87\rpm0.14} & \makebox{0.87\rpm0.14} & \makebox{0.87\rpm0.14} & \makebox{0.87\rpm0.14} & \makebox{0.87\rpm0.14} & \makebox{0.87\rpm0.14} & \makebox{0.87\rpm0.14} \\\midrule
{AMP-RNN} & \makebox{0.92\rpm0.05} & \makebox{0.87\rpm0.05} & \makebox{0.83\rpm0.11} & \makebox{0.84\rpm0.15} & \makebox{0.80\rpm0.15} & \makebox{0.77\rpm0.16} & \makebox{0.71\rpm0.18} & \makebox{0.62\rpm0.14} \\
{AMP-GRU} & \makebox{0.91\rpm0.07} & \makebox{0.90\rpm0.05} & \makebox{0.90\rpm0.09} & \makebox{0.87\rpm0.21} & \makebox{0.88\rpm0.21} & \makebox{0.88\rpm0.22} & \makebox{0.90\rpm0.19} & \makebox{0.89\rpm0.21} \\
{AMP-LSTM} & \makebox{0.91\rpm0.06} & \makebox{0.89\rpm0.04} & \makebox{0.88\rpm0.06} & \makebox{0.87\rpm0.10} & \makebox{0.81\rpm0.16} & \makebox{0.79\rpm0.20} & \makebox{0.77\rpm0.24} & \makebox{0.77\rpm0.24} \\\midrule
{AMP-ACT} & \makebox{\textbf{1.00\rpm0.00}} & \makebox{\textbf{0.99\rpm0.01}} & \makebox{\textbf{0.98\rpm0.02}} & \makebox{\textbf{0.99\rpm0.00}} & \makebox{\textbf{0.99\rpm0.00}} & \makebox{\textbf{0.99\rpm0.00}} & \makebox{\textbf{0.99\rpm0.00}} & \makebox{\textbf{0.99\rpm0.00}} \\
{AMP-Iter} & \makebox{\textbf{1.00\rpm0.00}} & \makebox{\textbf{1.00\rpm0.00}} & \makebox{\textbf{1.00\rpm0.00}} & \makebox{\textbf{1.00\rpm0.00}} & \makebox{\textbf{1.00\rpm0.00}} & \makebox{\textbf{1.00\rpm0.00}} & \makebox{\textbf{1.00\rpm0.00}} & \makebox{\textbf{1.00\rpm0.00}}\\\bottomrule
\end{tabular}}
\caption{Investigation of the oversmoothing effect. Table entries measure the accuracy of the subset of nodes that have a distance to the starting node that the model saw in the training set. Oversmoothing would mean that the introduction of nodes that are further away (and require additional rounds) causes a drop for the close-by nodes. A model that performs achieves the same result across all columns shows resilience against oversmoothing. All models except AMP-RNN demonstrate resilience. The only difference in accuracy can be explained away by the different performance on the training set.}
\label{tab:nonlocal_oversmoothing}
\end{table*}%

\begin{table*}[ht]
\centering
\resizebox{0.99\textwidth}{!}{
\begin{tabular}{@{}l*{8}{S[table-format=-3.4]}@{}}
\toprule
{Model} & {10} & {25} & {50} & {100} & {250} & {500} & {1000} & {2500}\\\midrule
{\multirow{2}{*}{NEG}} & \makebox{0.67\rpm0.12} & \makebox{0.54\rpm0.08} & \makebox{0.51\rpm0.03} & \makebox{0.51\rpm0.01} & \makebox{0.49\rpm0.01} & \makebox{0.50\rpm0.01} & \makebox{0.50\rpm0.00} & \makebox{0.50\rpm0.00}\\
 & \makebox{0.51\rpm0.02} & \makebox{0.49\rpm0.00} & \makebox{0.49\rpm0.00} & \makebox{0.49\rpm0.00} & \makebox{0.50\rpm0.00} & \makebox{0.50\rpm0.00} & \makebox{0.50\rpm0.00} & \makebox{0.50\rpm0.00}\\\midrule
{\multirow{2}{*}{Universal}} & \makebox{0.91\rpm0.05} & \makebox{0.74\rpm0.09} & \makebox{0.65\rpm0.11} & \makebox{0.61\rpm0.13} & \makebox{0.57\rpm0.14} & \makebox{0.56\rpm0.14} & \makebox{0.55\rpm0.14} & \makebox{0.55\rpm0.14}\\
 & \makebox{0.83\rpm0.01} & \makebox{0.67\rpm0.01} & \makebox{0.59\rpm0.00} & \makebox{0.54\rpm0.00} & \makebox{0.52\rpm0.00} & \makebox{0.51\rpm0.00} & \makebox{0.50\rpm0.00} & \makebox{0.50\rpm0.00}\\\midrule
{\multirow{2}{*}{IterGNN}} & \makebox{0.92\rpm0.10} & \makebox{0.82\rpm0.15} & \makebox{0.73\rpm0.13} & \makebox{0.65\rpm0.09} & \makebox{0.58\rpm0.05} & \makebox{0.55\rpm0.03} & \makebox{0.52\rpm0.01} & \makebox{0.51\rpm0.00}\\
 & \makebox{0.78\rpm0.08} & \makebox{0.66\rpm0.06} & \makebox{0.59\rpm0.05} & \makebox{0.55\rpm0.02} & \makebox{0.52\rpm0.01} & \makebox{0.51\rpm0.00} & \makebox{0.50\rpm0.00} & \makebox{0.50\rpm0.00}\\\midrule
{\multirow{2}{*}{AMP-RNN}} & \makebox{0.96\rpm0.02} & \makebox{0.85\rpm0.05} & \makebox{0.74\rpm0.08} & \makebox{0.69\rpm0.07} & \makebox{0.61\rpm0.09} & \makebox{0.58\rpm0.09} & \makebox{0.56\rpm0.05} & \makebox{0.54\rpm0.04}\\
 & \makebox{0.59\rpm0.03} & \makebox{0.56\rpm0.03} & \makebox{0.52\rpm0.03} & \makebox{0.51\rpm0.04} & \makebox{0.51\rpm0.03} & \makebox{0.50\rpm0.02} & \makebox{0.51\rpm0.01} & \makebox{0.50\rpm0.01}\\\midrule
{\multirow{2}{*}{AMP-GRU}} & \makebox{0.97\rpm0.01} & \makebox{0.89\rpm0.03} & \makebox{0.80\rpm0.05} & \makebox{0.73\rpm0.10} & \makebox{0.66\rpm0.07} & \makebox{0.63\rpm0.08} & \makebox{0.60\rpm0.05} & \makebox{0.56\rpm0.04}\\
 & \makebox{0.60\rpm0.02} & \makebox{0.58\rpm0.03} & \makebox{0.53\rpm0.03} & \makebox{0.51\rpm0.05} & \makebox{0.50\rpm0.03} & \makebox{0.50\rpm0.02} & \makebox{0.50\rpm0.01} & \makebox{0.51\rpm0.01}\\\midrule
{\multirow{2}{*}{AMP-LSTM}} & \makebox{0.97\rpm0.01} & \makebox{0.87\rpm0.03} & \makebox{0.77\rpm0.06} & \makebox{0.69\rpm0.10} & \makebox{0.61\rpm0.09} & \makebox{0.58\rpm0.10} & \makebox{0.56\rpm0.06} & \makebox{0.54\rpm0.05}\\
 & \makebox{0.60\rpm0.02} & \makebox{0.58\rpm0.03} & \makebox{0.52\rpm0.02} & \makebox{0.50\rpm0.03} & \makebox{0.50\rpm0.02} & \makebox{0.50\rpm0.02} & \makebox{0.50\rpm0.01} & \makebox{0.50\rpm0.00}\\\midrule
{\multirow{2}{*}{AMP-ACT}} & \makebox{1.00\rpm0.00} & \makebox{0.98\rpm0.02} & \makebox{0.96\rpm0.04} & \makebox{0.95\rpm0.05} & \makebox{0.93\rpm0.07} & \makebox{0.91\rpm0.10} & \makebox{0.91\rpm0.11} & \makebox{0.90\rpm0.12}\\
 & \makebox{\textbf{0.82\rpm0.02}} & \makebox{\textbf{0.75\rpm0.03}} & \makebox{\textbf{0.70\rpm0.02}} & \makebox{\textbf{0.68\rpm0.02}} & \makebox{\textbf{0.65\rpm0.02}} & \makebox{\textbf{0.63\rpm0.02}} & \makebox{\textbf{0.62\rpm0.03}} & \makebox{\textbf{0.59\rpm0.04}}\\\midrule
{\multirow{2}{*}{AMP-Iter}} & \makebox{1.0\rpm00.00} & \makebox{1.00\rpm0.00} & \makebox{1.00\rpm0.00} & \makebox{1.00\rpm0.00} & \makebox{1.00\rpm0.00} & \makebox{0.99\rpm5.99} & \makebox{0.99\rpm0.00} & \makebox{0.99\rpm0.00}\\
 & \makebox{\textbf{0.83\rpm0.03}} & \makebox{\textbf{0.76\rpm0.02}} & \makebox{\textbf{0.71\rpm0.01}} & \makebox{\textbf{0.68\rpm0.03}} & \makebox{\textbf{0.63\rpm0.04}} & \makebox{\textbf{0.61\rpm0.05}} & \makebox{0.58\rpm0.04} & \makebox{0.55\rpm0.06}\\\bottomrule
\end{tabular}}
\caption{Investigation for the oversquashing effect. The effect can be seen per method in the degradation from the first to the second row. The first row corresponds to the numbers in Table~\ref{tab:nonlocal} from the main body. The second row corresponds to the accuracy of solving three shortest path parity tasks at the same time. We only emphasize the best accuracy on this task. AMP-based approaches with self-supervised self termination perform best.}
\label{tab:nonlocal_oversquashing}
\end{table*}%

\end{document}